\documentclass[11pt]{article}
\usepackage[final]{acl}
\usepackage{times}
\usepackage{latexsym}
\usepackage[T1]{fontenc}
\usepackage[utf8]{inputenc}
\usepackage{microtype}
\usepackage{inconsolata}
\usepackage{graphicx}
\usepackage{booktabs}
\usepackage{amsmath}
\usepackage{xcolor}
\usepackage{algorithm}
\usepackage{algpseudocode}

\title{Margin-Guided Token Credit for\\Reinforcement Learning with Verifiable Rewards}

\author{Chi Liu \quad Xin Chen \quad Yuchuan Weng \quad Srinivasan Manoharan \\
  PayPal.AI}

\begin{document}
\maketitle
\begin{abstract}
In reinforcement learning with verifiable rewards, a language model is rewarded for a correct final answer, and GRPO passes that reward to every token of the solution in equal measure.
We instead place the credit where the model hesitated, in a method we call MCPO (Margin-Credit Policy Optimization). We measure hesitation by the top-2 margin $M=p_1-p_2$ of the policy's next-token distribution, map it to a score with two fixed thresholds, and spread each solution's advantage over its tokens so that the mean stays unchanged. The margin is read from a forward pass the trainer already performs, so the method needs no extra model evaluation and no batch statistics.
We train MiniCPM5-2B-Base with LoRA for 200 steps on 2,048 DAPO-Math problems and evaluate on six competition-math benchmarks. MCPO beats an unmodified GRPO baseline for all five data seeds, by 6.55 points of macro avg@32 on average (95\% CI $[4.1, 9.0]$), and a Qwen3-4B-Base replication improves for all three seeds ($+4.75$).
The gain is largest on problems that GRPO solves only some of the time and grows with $k$ in pass@$k$, reaching 13.8 points at pass@32. MCPO models also write two to three times longer answers, and runs that write longer score higher; our experiments do not separate the two effects.

\end{abstract}

\section{Introduction}
\begin{figure*}[t]
\centering
\includegraphics[width=\textwidth]{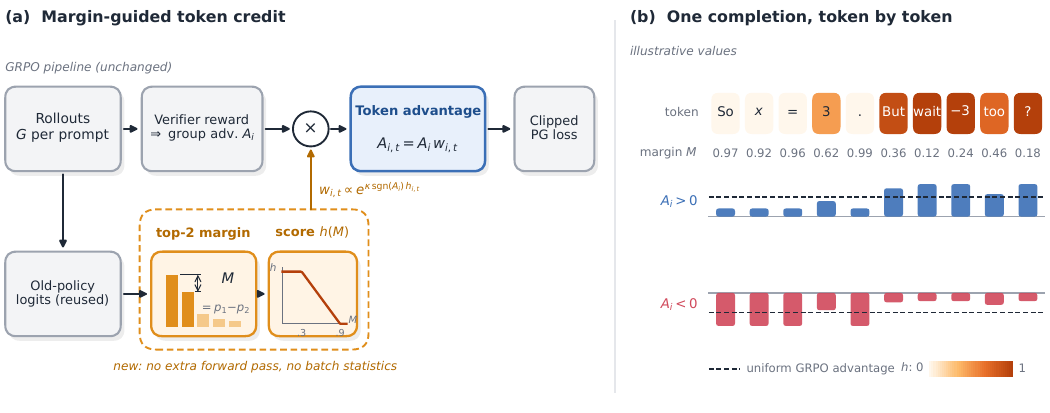}
\caption{Overview. (a) Relative to the GRPO baseline, the method changes only the token-level advantages; the new components are orange. The old-policy forward pass that the trainer already runs also gives the top-2 margin $M=p_1-p_2$ at every position, and a fixed-threshold map turns it into a score $h\in[0,1]$. (b) An illustrative completion. Tokens with a clear winner get $h=0$ and close calls get $h\approx1$. In a positive completion the close calls receive more of the advantage; in a negative one they receive less of the penalty. The mean token advantage stays $A_i$.}
\label{fig:overview}
\end{figure*}

Reinforcement learning with verifiable rewards (RLVR) has become the standard recipe for teaching language models to reason~\citep{lambert2024tulu,guo2025r1}. GRPO~\citep{shao2024deepseekmath}, its most common form, scores each sampled solution against the other solutions to the same problem and hands the resulting advantage to every token of the solution. A 3{,}000-token derivation thus credits ``Let'' and ``='' exactly as much as the step that decides the argument.

Most of those tokens were never in doubt. A small fraction of positions, where the model weighs two or more continuations, decide where a derivation goes~\citep{wang2025beyond}. Recent work locates these positions with token entropy: it trains only on the highest-entropy tokens~\citep{wang2025beyond}, adds an entropy bonus to the advantage~\citep{cheng2025exploration}, weights token rewards by entropy~\citep{tan2025gtpo}, or constrains high- and low-entropy tokens differently~\citep{liu2025archer}. Entropy, however, measures the spread of the whole distribution, not the closeness of the decision. A position with one clear favorite and a long flat tail has high entropy although the next token is not in question. And because raw entropies are hard to compare, most of these methods normalize them against the batch or the group, so the same distribution can earn different credit depending on what else was sampled.

We ask a narrower question at each position: how close was the call between the two leading candidates? The answer is the top-2 margin $M=p_1-p_2$. We map it to a score with two fixed thresholds ($M\ge0.9$ gives 0, $M\le0.3$ gives 1) and use the score to redistribute each completion's advantage over its tokens, keeping the mean unchanged (Figure~\ref{fig:overview}). Close calls get more of a reward and less of a penalty. The score depends on one distribution only, and it comes from the old-policy forward pass the trainer already runs, so it adds no model evaluation. We call the method MCPO, for Margin-Credit Policy Optimization.

We compare MCPO with an unmodified GRPO baseline over 16 training runs on two base models and evaluate on six competition-math benchmarks. MCPO improves macro avg@32 in all eight (model, data seed) pairs, by $+6.55$ points on MiniCPM5-2B-Base (95\% CI $[4.1, 9.0]$) and $+4.75$ on Qwen3-4B-Base. The gain is concentrated on problems of intermediate difficulty and grows with the sampling budget, reaching $+13.8$ points at pass@32. It comes with answers two to three times longer, and across runs accuracy tracks length closely (Sections~\ref{sec:length} and~\ref{sec:lenacc}).

Our contributions are:
\begin{itemize}
  \item a token-credit rule driven by a fixed-threshold top-2 margin, which needs no extra forward pass and no batch statistics (Section~\ref{sec:method});
  \item a controlled comparison with GRPO across 16 runs, with seed-level and problem-level uncertainty estimates (Section~\ref{sec:results});
  \item an analysis of where the gain appears and of its link to answer length (Sections~\ref{sec:difficulty}--\ref{sec:passk}).
\end{itemize}

\section{Method}
\label{sec:method}
\paragraph{GRPO.}
For a prompt $q$, the policy $\pi_\theta$ samples $G$ completions $o_1,\dots,o_G$ and a verifier assigns each a reward $r_i$. GRPO~\citep{shao2024deepseekmath} standardizes the rewards within the group, $A_i=(r_i-\bar r)/(\sigma_r+\epsilon)$, and minimizes a clipped surrogate loss:
\begin{multline}
\mathcal L=-\frac1{\sum_i|o_i|}\sum_{i}\sum_{t}\min\!\big(\rho_{i,t}A_{i,t},\\
\mathrm{clip}(\rho_{i,t},1\!\pm\!\varepsilon)A_{i,t}\big),
\end{multline}
where $\rho_{i,t}$ is the importance ratio of token $t$. The codebase we build on normalizes over tokens as written, clips asymmetrically ($\varepsilon=0.2$ below and $0.28$ above), uses no KL term, and caps the importance weight that corrects for the vLLM sampler at 2. Plain GRPO sets $A_{i,t}=A_i$ for every token. We replace this constant with a token-dependent $A_{i,t}$ whose mean over the completion is still $A_i$, and keep sampling, clipping and length handling unchanged.

\paragraph{Spreading the advantage over tokens.}
Given a score $h_{i,t}\in[0,1]$ for each token, where large means uncertain, the token advantage is
\begin{equation}
A_{i,t}=A_i\,\frac{e^{\kappa\,\mathrm{sign}(A_i)\,h_{i,t}}}{\frac1{|T_i|}\sum_{t'\in T_i}e^{\kappa\,\mathrm{sign}(A_i)\,h_{i,t'}}},
\label{eq:adv}
\end{equation}
where $T_i$ is the set of valid tokens of completion $i$. Dividing by the mean weight keeps the mean token advantage at $A_i$. In a positive-advantage completion, uncertain tokens receive more of the advantage; in a negative-advantage one they receive less, so confident tokens carry more of the penalty. (Negative advantage is not the same as an incorrect answer, because the reward also includes the overlong penalty.) We use $\kappa=\ln4$ for both signs. This weighting rule is due to \citet{yeo2026surprising}, who pair it with token entropy normalized by the 10\% and 90\% quantiles of the rollout batch; we build on their public code, switch that entropy score off, and replace it with the margin score below.

\begin{figure*}[t]
\centering
\includegraphics[width=\textwidth]{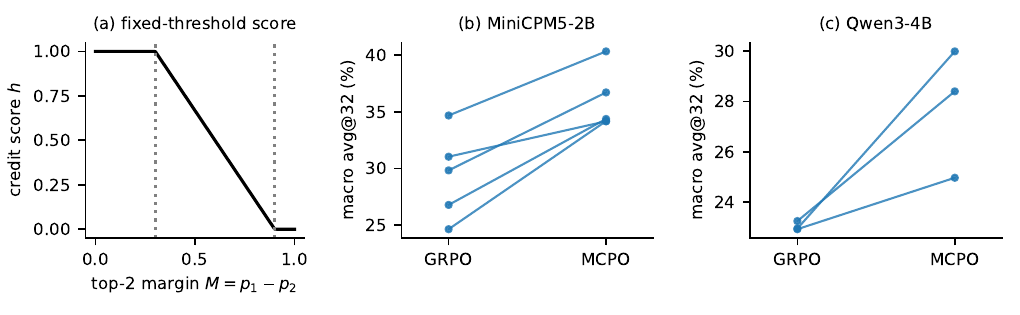}
\caption{(a) The fixed-threshold score $h(M)$ of Equation~\ref{eq:h}. (b,c) Macro avg@32 over six benchmarks per data seed, GRPO vs.\ MCPO, for MiniCPM5-2B-Base (5 seeds) and Qwen3-4B-Base (3 seeds). Each line is one data seed.}
\label{fig:main}
\end{figure*}

\paragraph{The margin score.}
Let $p$ be the rollout policy's next-token distribution at a position, after temperature scaling, and let $p_1\ge p_2$ be its two largest probabilities. We set
\begin{equation}
M=p_1-p_2,\; h=\mathrm{clip}\!\left(\tfrac{(1-M)-\ell}{u-\ell},0,1\right),
\label{eq:h}
\end{equation}
with $\ell=0.1$ and $u=0.7$. A margin of 0.9 or more gives $h=0$, a margin of 0.3 or less gives $h=1$, and $h$ is linear in between (Figure~\ref{fig:main}a). The thresholds are absolute, so $h$ depends on one token's distribution and on nothing else in the batch. (The normalized weight in Equation~\ref{eq:adv} still depends on the other tokens of the same completion, and $A_i$ on the rewards of its group.)

\paragraph{Implementation.}
To correct for the vLLM sampler, the trainer already runs a forward pass of the old policy over each rollout. We wrap the log-softmax routine for that one call so that it also records the top two probabilities at every position, and compute Equation~\ref{eq:adv} from them (Algorithm~\ref{alg:margin}). No extra model pass is needed. The margin inherits the bf16 precision of the logits: on random bf16 logits over a 130{,}560-token vocabulary at temperature 1, the largest deviation from fp32 we observed was about $2\times10^{-3}$ (about ten times larger at temperature 0.7; training samples at temperature 1.0). For the second mini-batch of a rollout the margin is one optimizer step old.

\begin{algorithm}[t]
\small
\caption{MCPO for one rollout (changes to a GRPO trainer in blue).}
\label{alg:margin}
\begin{algorithmic}[1]
\State sample $G$ completions per prompt with vLLM; compute rewards $r_i$ and group advantages $A_i$
\State old-policy forward pass over the rollout $\to$ logits (already run by the trainer)
\State \textcolor{blue}{$M_{i,t}\gets p_1-p_2$ from these logits (temperature-scaled)}
\State \textcolor{blue}{$h_{i,t}\gets\mathrm{clip}\big((1-M_{i,t}-\ell)/(u-\ell),0,1\big)$}
\For{each completion $i$}
  \State \textcolor{blue}{$w_{i,t}\gets e^{\kappa\,\mathrm{sign}(A_i)h_{i,t}}$; \; $A_{i,t}\gets A_i\,w_{i,t}/\overline{w_i}$}
\EndFor
\State two optimizer steps on mini-batches of the rollout with the unchanged clipped loss using $A_{i,t}$
\end{algorithmic}
\end{algorithm}

\subsection{Properties and design choices}
\label{sec:props}
\paragraph{What the reweighting preserves.}
Every completion keeps $\frac1{|T_i|}\sum_t A_{i,t}=A_i$. Its share of the gradient can still change, because per-token gradients differ, the loss is clipped, and the loss normalizes by the total number of tokens in the batch. Within one completion, two tokens' raw weights differ by at most a factor $e^{\kappa}=4$. A completion whose tokens share one score gets uniform advantages, and a completion with $A_i=0$ stays at zero.

\begin{figure}[t]
\centering
\includegraphics[width=\columnwidth]{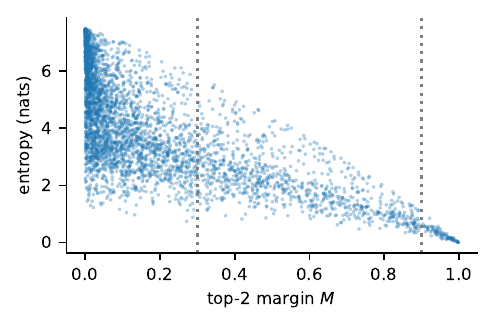}
\caption{Entropy vs.\ top-2 margin for 4{,}000 \emph{synthetic} next-token distributions (random Gaussian logits over 2{,}000 tokens with 0--3 boosted candidates; not model outputs). Dotted lines: the thresholds $0.3$ and $0.9$. For $M\in[0.3,0.4)$ the 5th--95th percentile of the entropy is 1.5--4.8 nats.}
\label{fig:entropy}
\end{figure}

\paragraph{Why a margin.}
Entropy summarizes the whole distribution, including its tail. A position with a clear favorite and a long flat tail can have sizeable entropy even though the next token is not in question. The margin asks only whether the favorite has a close rival. On synthetic logits, positions with the same margin differ by several nats of entropy (Figure~\ref{fig:entropy}). Using the gap between the two leading candidates to measure uncertainty is an old idea; it is the margin criterion of uncertainty sampling in active learning~\citep{settles2009active}. We do not claim that the margin finds reasoning forks better than entropy does: our experiments compare MCPO only with plain GRPO.

\paragraph{Why fixed thresholds.}
Under quantile normalization, the same distribution can score $h=0$ in a batch of uncertain tokens and $h=1$ in a batch of confident ones. Under absolute thresholds, a margin of 0.9 always scores 0 and a margin of 0.3 always scores 1. The price is two hyperparameters that do not adapt to the model or the stage of training; we use one untuned setting throughout.

\paragraph{A worked example.}
Consider a positive completion with three tokens whose margins are $(0.98,\,0.60,\,0.20)$. Their scores are $(0,\,0.5,\,1)$ and their raw weights $(1,\,2,\,4)$, with mean $7/3$, so the token advantages are $A_i\cdot(0.43,\,0.86,\,1.71)$. In a negative completion the same margins give raw weights $(1,\,0.5,\,0.25)$ with mean $7/12$, and token advantages $A_i\cdot(1.71,\,0.86,\,0.43)$ with $A_i<0$. The confident token now takes the largest share of the penalty, and the close call the smallest.

\paragraph{Cost.}
The method adds no model evaluation and no rollout. It adds a top-2 selection over the vocabulary, a log-softmax over the vocabulary gathered at those two indices, the weighting of Equation~\ref{eq:adv}, and the memory to hold the margins until the advantages are computed.

\section{Experiments}
\label{sec:exp}
\paragraph{Training.}
We build on the public training code of \citet{yeo2026surprising} at a pinned commit and train with LoRA~\citep{hu2022lora} (rank 32, $\alpha=64$, all linear layers). Each run uses 2{,}048 problems sampled from DAPO-Math-17k~\citep{yu2025dapo}; the \emph{data seed} decides which problems. The TRL training seed stays at its default, so our seeds vary the training problems but not the optimization noise that is known to make RL results fragile~\citep{henderson2018matters,agarwal2021precipice}, and the intervals below are conditional on that. A rollout draws 128 completions (16 prompts $\times$ 8) at temperature 1.0 and feeds two optimizer steps of 64 completions each. We train for 200 optimizer steps, i.e.\ 100 rollouts, with learning rate $10^{-5}$ and completions of at most 10{,}240 tokens. The reward is answer correctness, graded by Math-Verify~\citep{mathverify}, plus the DAPO soft overlong penalty~\citep{yu2025dapo}, which starts at 8{,}192 tokens. Rollouts are generated by vLLM~\citep{kwon2023vllm} on the training GPU. The baseline is the codebase's GRPO script, unmodified. The margin runs use the codebase's token-credit trainer, a subclass of the same TRL GRPO trainer, with its entropy credit set to zero and the margin score added; both entry points read every training argument from one shared defaults table, so the two arms share model, data, seed, step count and all optimization settings (Table~\ref{tab:settings}). Our CPU tests check that, apart from the advantages, the rollout tensors equal those of the codebase's credit trainer with entropy credit off, but we did not compare against the plain GRPO trainer or train a zero-strength run, so a residual difference between the two training paths is not excluded (Section~\ref{sec:discussion}). Our main model is MiniCPM5-2B-Base~\citep{hu2024minicpm} with five data seeds (43, 45, 100, 1024, 31415). We repeat the comparison on Qwen3-4B-Base~\citep{yang2025qwen3} with three data seeds (42, 43, 45) and the same recipe.

\paragraph{Evaluation.}
We run the codebase's evaluation script unchanged on AIME24, AIME25, AIME26, HMMT26, AMC23 and MATH500 level~5~\citep{hendrycks2021math,lightman2023verify}, 297 problems in all.\footnote{The competition sets are the official AIME, HMMT (February) and AMC problems as distributed with the codebase's evaluation data. AIME26 and HMMT26 were held in February 2026, after the release of Qwen3-4B-Base.} Each problem gets 32 samples at temperature 0.7, top-$p$ 0.8 and top-$k$ 20, with up to 16{,}384 new tokens. For one benchmark, avg@32 is the accuracy averaged over all 32 samples of all its problems; it estimates single-sample accuracy, not the chance of success within 32 attempts. Macro avg@32 averages the six benchmarks. A seed's gain is the macro avg@32 of its margin run minus that of its GRPO run. We report the mean gain over seeds, a two-level bootstrap 95\% interval (seeds resampled with replacement; within each benchmark, problems resampled with the same indices for every seed; 20{,}000 draws), and a two-sided one-sample $t$-test of the per-seed gains against zero.

\begin{table}[t]
\centering\small
\begin{tabular}{ll}
\toprule
\textbf{Setting} & \textbf{Value}\\
\midrule
MCPO & $\kappa=\ln4$, $\ell=0.1$, $u=0.7$\\
Adapter & LoRA $r{=}32$, $\alpha{=}64$\\
Learning rate / steps & $10^{-5}$ / 200\\
Rollout & 16 prompts $\times$ 8 samples\\
Steps per rollout & 2 (64 samples each)\\
Max.\ completion & 10{,}240 tokens\\
Training set & 2{,}048 DAPO-Math problems\\
Evaluation & 297 problems, 32 samples each\\
Eval.\ sampling & $T{=}0.7$, top-$p$ 0.8, top-$k$ 20\\
Eval.\ max.\ length & 16{,}384 tokens\\
\bottomrule
\end{tabular}
\caption{Settings shared by the GRPO and MCPO runs.}
\label{tab:settings}
\end{table}

\subsection{Main results}
\label{sec:results}
\begin{table}[t]
\centering\small
\begin{tabular}{lrrr}
\toprule
\textbf{Data seed} & \textbf{GRPO} & \textbf{Margin} & \textbf{Gain}\\
\midrule
43 & 31.04 & 34.16 & +3.12\\
45 & 29.83 & 36.71 & +6.87\\
100 & 34.67 & 40.32 & +5.65\\
1024 & 26.79 & 34.37 & +7.57\\
31415 & 24.65 & 34.18 & +9.53\\
\midrule
Mean & 29.40 & 35.95 & +6.55\\
\bottomrule
\end{tabular}
\caption{MiniCPM5-2B-Base: macro avg@32 (\%) over six benchmarks. Bootstrap 95\% CI of the mean gain $[+4.14,+8.98]$; $t$-test $p=0.004$ ($df=4$).}
\label{tab:minicpm}
\end{table}

\begin{table}[t]
\centering\small
\begin{tabular}{lrrr}
\toprule
\textbf{Data seed} & \textbf{GRPO} & \textbf{Margin} & \textbf{Gain}\\
\midrule
42 & 22.93 & 24.97 & +2.04\\
43 & 22.95 & 30.00 & +7.05\\
45 & 23.25 & 28.41 & +5.16\\
\midrule
Mean & 23.04 & 27.79 & +4.75\\
\bottomrule
\end{tabular}
\caption{Qwen3-4B-Base: macro avg@32 (\%). Bootstrap 95\% CI $[+2.06,+7.65]$; $t$-test $p=0.083$ ($df=2$; three seeds, so the $t$-test has little power).}
\label{tab:qwen}
\end{table}

MCPO beats GRPO for every data seed (Tables~\ref{tab:minicpm} and~\ref{tab:qwen}; Figure~\ref{fig:main}b,c). On MiniCPM5-2B-Base the mean gain is $+6.55$ points, 22\% above GRPO's 29.40, and the bootstrap interval excludes zero. On Qwen3-4B-Base the mean gain is $+4.75$. Its bootstrap interval excludes zero, but three seeds give only ten distinct unordered resamples (27 ordered draws), so it rests on little seed-level information; the $t$-test gives $p=0.083$. We take the Qwen result as consistent in sign, not as established.

\begin{table*}[t]
\centering\small
\begin{tabular}{lrrrrrrrrr}
\toprule
& & \multicolumn{4}{c}{\textbf{MiniCPM5-2B-Base}} & \multicolumn{4}{c}{\textbf{Qwen3-4B-Base}}\\
\cmidrule(lr){3-6}\cmidrule(lr){7-10}
\textbf{Benchmark} & $n$ & GRPO & Margin & Gain & 95\% CI & GRPO & Margin & Gain & 95\% CI\\
\midrule
AIME24 & 30 & 15.77 & 27.90 & +12.13 & [+8.4, +16.2] & 11.94 & 16.63 & +4.69 & [+1.7, +8.2]\\
AIME25 & 30 & 15.96 & 21.10 & +5.15 & [+2.3, +8.4] & 8.16 & 14.13 & +5.97 & [+2.1, +10.7]\\
AIME26 & 30 & 14.58 & 21.23 & +6.65 & [+3.6, +10.2] & 7.19 & 13.23 & +6.04 & [+2.3, +10.6]\\
HMMT26 & 33 & 12.01 & 14.98 & +2.97 & [$-$1.2, +7.1] & 5.78 & 8.96 & +3.19 & [0.0, +7.0]\\
AMC23 & 40 & 59.59 & 64.80 & +5.20 & [+2.5, +8.3] & 53.67 & 55.57 & +1.90 & [$-$1.6, +5.5]\\
MATH500 L5 & 134 & 58.47 & 65.67 & +7.20 & [+5.1, +9.4] & 51.51 & 58.22 & +6.71 & [+4.2, +9.4]\\
\midrule
Macro avg. & 297 & 29.40 & 35.95 & +6.55 & [+4.1, +9.0] & 23.04 & 27.79 & +4.75 & [+2.1, +7.7]\\
\bottomrule
\end{tabular}
\caption{avg@32 (\%) per benchmark, averaged over data seeds. The per-benchmark intervals are bootstrap 95\% intervals over \emph{problems} for the seed-averaged gain (10{,}000 draws) and therefore do not include seed variation; the macro-average intervals are the two-level bootstrap intervals of Tables~\ref{tab:minicpm} and~\ref{tab:qwen}.}
\label{tab:bench}
\end{table*}

\paragraph{Per benchmark.}
On MiniCPM the gain averaged over seeds is positive on every benchmark (Table~\ref{tab:bench}), from $+2.97$ on HMMT26, where GRPO is weakest, to $+12.13$ on AIME24. Problem-level intervals exclude zero on five of six benchmarks for MiniCPM (not on HMMT26, with 33 problems) and on four of six for Qwen (the HMMT26 interval touches zero and the AMC23 interval contains it). All 48 per-seed, per-benchmark gains are positive (Appendix~\ref{app:bench}). Macro pass@32 rises over GRPO for every seed as well, by $6.0$--$20.3$ points on MiniCPM and $3.8$--$16.7$ on Qwen: more problems are solved at least once, although problems GRPO never solves gain only about 2 points (Table~\ref{tab:diff}).

\subsection{Gains by problem difficulty}
\label{sec:difficulty}
To see which problems gain, we bin them by GRPO accuracy and compare the methods within each bin (Table~\ref{tab:diff}). Binning on the same samples that are then compared would build in regression to the mean, so we split each problem's 32 samples: the bins use GRPO's first 16 samples (pooled over a model's data seeds), and all accuracies in the table use the other 16. The gain sits on problems that GRPO solves some of the time: $+12.5$ and $+11.6$ points in the two middle bins for MiniCPM, $+11.7$ and $+10.8$ for Qwen. Problems that GRPO solves in more than 75\% of samples barely move, and problems it never solves gain about 2 points.

\begin{table}[t]
\centering\small
\setlength{\tabcolsep}{4pt}
\begin{tabular}{lrrrr}
\toprule
\textbf{GRPO acc.} & $n$ & \textbf{GRPO} & \textbf{Margin} & \textbf{Diff}\\
\midrule
\multicolumn{5}{l}{\emph{MiniCPM5-2B-Base}}\\
0 & 71 & 0.3 & 2.9 & +2.6\\
$(0,25\%]$ & 68 & 8.8 & 21.3 & +12.5\\
$(25\%,75\%]$ & 72 & 48.4 & 60.1 & +11.6\\
$>75\%$ & 86 & 92.2 & 93.3 & +1.1\\
\midrule
\multicolumn{5}{l}{\emph{Qwen3-4B-Base}}\\
0 & 102 & 0.3 & 2.4 & +2.1\\
$(0,25\%]$ & 71 & 8.7 & 20.3 & +11.7\\
$(25\%,75\%]$ & 51 & 49.6 & 60.3 & +10.8\\
$>75\%$ & 73 & 93.5 & 93.5 & $-$0.1\\
\bottomrule
\end{tabular}
\caption{Accuracy (\%) by difficulty bin. Bins use GRPO's first 16 samples per problem, pooled over data seeds; all accuracies use the remaining 16 samples (held out from binning). Each problem counts equally.}
\label{tab:diff}
\end{table}

\subsection{Answer length}
\label{sec:length}
\begin{figure}[t]
\centering
\includegraphics[width=\columnwidth]{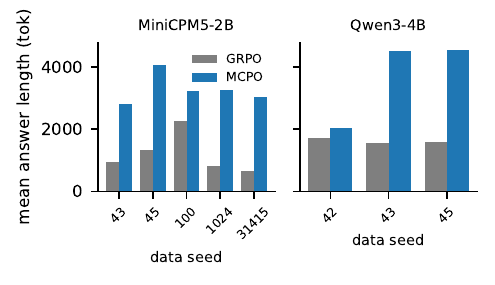}
\caption{Mean answer length over all evaluation answers, per data seed.}
\label{fig:length}
\end{figure}

MCPO models write longer answers (Figure~\ref{fig:length}). Over all evaluation answers, the MiniCPM GRPO models average 652--2{,}264 tokens per seed (mean 1{,}208), the MCPO models 2{,}812--4{,}066 (mean 3{,}275). On Qwen3-4B-Base the means are 1{,}621 and 3{,}697 tokens. The ratio is largest on the hardest benchmarks and smallest on MATH500 level~5 (Table~\ref{tab:len}). Across seeds, the gain grows with the length ratio: $r=0.65$ over the five MiniCPM seeds and $r=0.92$ over the three Qwen seeds, neither significant at these sample sizes. Two GRPO baselines are strikingly short (MiniCPM seeds 31415 and 1024, 652 and 831 tokens against a 10{,}240-token training budget), and they show the two largest gains. Part of the improvement may therefore come from avoiding the very short answers that some GRPO runs produce in this short LoRA regime.

\begin{table}[t]
\centering\small
\setlength{\tabcolsep}{3pt}
\begin{tabular}{lrrrr}
\toprule
& \multicolumn{2}{c}{\textbf{MiniCPM}} & \multicolumn{2}{c}{\textbf{Qwen3-4B}}\\
\textbf{Benchmark} & GRPO & Margin & GRPO & Margin\\
\midrule
AIME24 & 1{,}511 & 4{,}785 & 2{,}704 & 5{,}887\\
AIME25 & 1{,}404 & 4{,}364 & 1{,}984 & 4{,}596\\
AIME26 & 1{,}364 & 4{,}314 & 2{,}142 & 5{,}473\\
HMMT26 & 1{,}329 & 4{,}922 & 2{,}054 & 5{,}463\\
AMC23 & 1{,}272 & 2{,}911 & 1{,}444 & 2{,}773\\
MATH500 L5 & 1{,}011 & 2{,}163 & 1{,}127 & 2{,}450\\
\bottomrule
\end{tabular}
\caption{Mean answer length in tokens per benchmark, averaged over data seeds (mean of per-benchmark means).}
\label{tab:len}
\end{table}

Our experiments do not identify how much of the gain flows through longer reasoning (see the hypothesis in Section~\ref{sec:discussion}). A length-matched or longer-trained GRPO baseline would be needed to separate the credit signal from the length effect.

\subsection{Accuracy and length across runs}
\label{sec:lenacc}
\begin{figure}[t]
\centering
\includegraphics[width=\columnwidth]{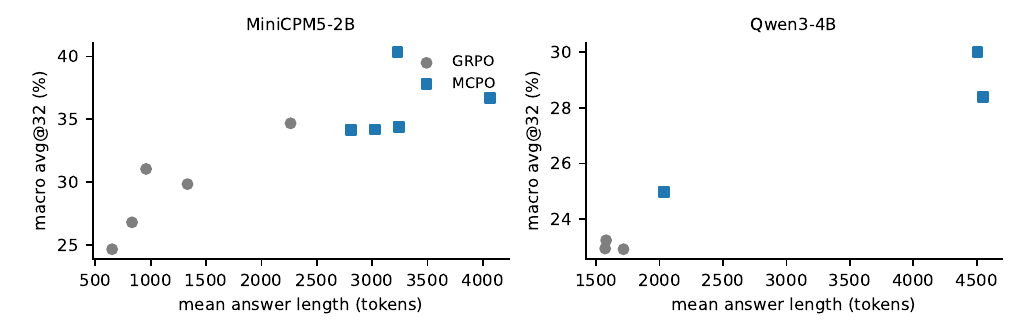}
\caption{Macro avg@32 vs.\ mean answer length for every run (one point per method and data seed).}
\label{fig:lenacc}
\end{figure}

Plotting each run's macro avg@32 against its mean answer length (Figure~\ref{fig:lenacc}) gives $r=0.87$ over the ten MiniCPM runs and $r=0.97$ over the six Qwen runs: runs that write longer score higher, and the MCPO runs sit at the long end. Within a method there are too few runs to say much (Appendix~\ref{app:within}). Across MiniCPM seeds the gain also falls as the GRPO score rises ($r=-0.73$), but this is partly built in, since the gain contains the GRPO score with a minus sign. With 3--5 runs per method, these correlations cannot tell whether MCPO raises accuracy by lengthening reasoning, whether longer reasoning follows from better training, or whether one change in the policy drives both.

\subsection{Sampling budget}
\label{sec:passk}
\begin{figure*}[t]
\centering
\includegraphics[width=\textwidth]{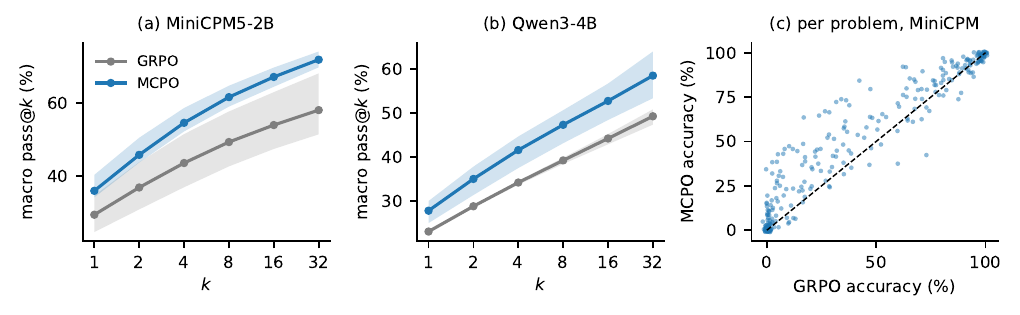}
\caption{(a,b) Macro pass@$k$ over the six benchmarks (mean over data seeds; band: min--max over seeds). (c) Per-problem accuracy, GRPO vs.\ MCPO, MiniCPM5-2B-Base, pooled over seeds (small jitter added); points above the dashed diagonal favor MCPO.}
\label{fig:analysis}
\end{figure*}

We estimate pass@$k$ with the unbiased estimator of \citet{chen2021codex} from the 32 samples per problem, macro-average over benchmarks and average over seeds (Figure~\ref{fig:analysis}a,b). At $k=32$ the estimator reduces to the empirical pass@32. MCPO leads at every $k$ from 1 to 32 on both models, and the lead grows with $k$: on MiniCPM from $+6.5$ points at $k=1$ (29.4 to 35.9) to $+13.8$ at $k=32$ (58.1 to 71.9), on Qwen from $+4.7$ to $+9.3$ (49.3 to 58.6 at $k=32$). Per problem, most points lie above the diagonal (Figure~\ref{fig:analysis}c). Pooled over seeds, MCPO is more accurate than GRPO on 189 of 297 MiniCPM problems and less accurate on 57 (Qwen: 145 and 62).

\subsection{Discussion}
\label{sec:discussion}
\paragraph{Supported claims.}
At an equal number of optimizer steps (not an equal number of generated tokens), MCPO models beat GRPO models in all eight (model, data seed) pairs, on every benchmark when averaged over seeds, and at every sampling budget up to 32. The gain is largest on problems of intermediate difficulty. For MiniCPM5-2B-Base the seed-level interval excludes zero.

\paragraph{Open questions.}
The results do not show that the margin is a better score than entropy, because we ran neither the entropy score of \citet{yeo2026surprising} nor any other entropy-based credit, or that it matters \emph{which} tokens get the credit, because we ran no random-score control. They cannot rule out that longer answers explain part of the gain. And whether the advantage over GRPO survives longer training, a tuned GRPO baseline or full fine-tuning is unknown. Section~\ref{sec:conclusion} names the runs that would answer these questions.

\paragraph{A hypothesis.}
The sign-aware weighting suggests why MCPO might keep answers longer. In a negative-advantage completion it moves the penalty away from close calls and onto confident tokens, so the steps where a derivation branches are punished less; in a positive one it rewards those steps more. If close calls often open new reasoning steps, uniform credit would prune them and shorten answers, and MCPO would keep them. The training runs log the share of uncertain tokens in positive and negative completions, which is where a test would start; we have not analyzed those logs.

\paragraph{Reproducibility.}
All runs use the codebase's data pipeline, reward functions and evaluation. The supplementary code pins the codebase and provides the training commands, checksums of the sampled training sets, and CPU tests of the margin computation. It documents the MiniCPM recipe in detail; the Qwen3-4B-Base runs use the same recipe with a different base model. One training run uses one 80\,GB GPU.

\section{Related Work}
\paragraph{RL with verifiable rewards.}
Training on rewards that a program can check, such as agreement of a final answer with a reference, goes back to self-training on correct rationales~\citep{zelikman2022star} and is now the standard recipe for reasoning models~\citep{lambert2024tulu,guo2025r1,kimi2025k15,yang2024qwen25math}. Most open recipes drop the learned critic of PPO~\citep{schulman2017ppo,schulman2016gae} in favor of baselines built from several samples of the same prompt: RLOO~\citep{ahmadian2024rloo}, ReMax~\citep{li2024remax}, REINFORCE++~\citep{hu2025reinforcepp} and GRPO~\citep{shao2024deepseekmath}. Later work revises the GRPO objective and its sampling, through decoupled clipping and dynamic sampling~\citep{yu2025dapo}, the removal of length and difficulty biases~\citep{liu2025drgrpo}, sequence-level importance ratios~\citep{zheng2025gspo}, or a value model reintroduced for long chains~\citep{yue2025vapo}, and studies how to scale or prolong training from a base model~\citep{hu2025orz,zeng2025simplerl,liu2025prorl}. Whether RLVR teaches new reasoning or mainly sharpens what the base model can already do is debated~\citep{yue2025does,shao2025spurious}, and objectives that target pass@$k$ directly have been proposed in response~\citep{chen2025passk}. Reported gains are also sensitive to evaluation seeds and settings~\citep{hochlehnert2025sober} and, for some base models, to contamination of the benchmarks~\citep{wu2025reasoning}. In nearly all of these methods, one advantage covers the whole completion.

\paragraph{Credit below the sequence level.}
An outcome reward says little about which step earned it. Process reward models score intermediate steps, trained on human labels~\citep{lightman2023verify} or on automatically estimated step values~\citep{wang2024mathshepherd,setlur2024rewarding}; PRIME derives implicit process rewards from an outcome-trained model~\citep{cui2025prime}. VinePPO~\citep{kazemnejad2024vineppo} and segment-level methods~\citep{guo2025spo} estimate values by Monte Carlo resampling from intermediate states, and contrastive estimation can locate critical tokens that lead to wrong answers~\citep{lin2024critical}; KTAE estimates token-level advantages from rollout statistics without an extra model~\citep{sun2025ktae}. These methods buy finer credit with extra models, labels or samples. MCPO uses none of them: it reuses the policy's own next-token distribution.

\paragraph{Entropy as a credit signal.}
\citet{cui2025entropy} trace how policy entropy collapses during RL for reasoning, and \citet{agarwal2025entmin} show that minimizing entropy alone can improve reasoning. \citet{wang2025beyond} find that a minority of high-entropy tokens act as forks in reasoning chains, and that restricting the policy gradient to the top 20\% of tokens by entropy preserves or improves accuracy. \citet{cheng2025exploration} add an entropy term to the advantage; GTPO~\citep{tan2025gtpo} redistributes rewards over tokens by entropy ratios; Archer~\citep{liu2025archer} constrains high- and low-entropy tokens differently. \citet{yeo2026surprising} treat success and failure asymmetrically: they spread the advantage with Equation~\ref{eq:adv} and score tokens by quantile-normalized entropy. All of these rely on entropy, and most make it comparable through a batch- or group-relative step. \citet{yang2025lowprob} show that low-probability tokens dominate the GRPO gradient and reduce their influence by reweighting. \citet{bigelow2024forking} find forking tokens directly, by costly resampling. We keep the weighting rule of \citet{yeo2026surprising} and replace its score with a top-2 margin under absolute thresholds.

\paragraph{Confidence and uncertainty.}
Language models carry usable information about their own correctness~\citep{kadavath2022know}, although modern networks are often miscalibrated~\citep{guo2017calibration}, and uncertainty over generations is better measured over meanings than over strings~\citep{kuhn2023semantic}. The gap between the two most probable labels is the margin criterion of uncertainty sampling in active learning~\citep{scheffer2001active,settles2009active}. In LLM reasoning, CoT-decoding~\citep{wang2024cotdecoding} ranks decoding paths by the average top-1 minus top-2 probability gap over the answer tokens, and Intuitor~\citep{zhao2025intuitor} trains with GRPO on the model's self-certainty in place of an external reward. Self-certainty also ranks samples for best-of-$N$~\citep{kang2025selfcertainty}, token confidence filters reasoning traces at test time~\citep{fu2025deepconf}, and confidence alone can serve as an RL reward~\citep{prabhudesai2025rent}. These works use confidence to choose paths or to define the reward. We keep the verifiable reward and use the token-level margin only to decide where within a completion its advantage goes.

\paragraph{Length of reasoning.}
Chains of thought~\citep{wei2022cot,wang2023selfconsistency} grow during RL training, and how long they grow depends on the reward shaping and the training recipe~\citep{yeo2025demystifying}. Other work trains models to reason more concisely~\citep{fatemi2025concise} or to follow a length budget~\citep{aggarwal2025l1}. Our MCPO runs write longer answers than their GRPO counterparts; Sections~\ref{sec:length} and~\ref{sec:lenacc} discuss what this means for our comparison.

\section{Conclusion}
\label{sec:conclusion}
A top-2 margin with two fixed thresholds, read from a forward pass the trainer already runs, improves short GRPO training: MCPO beat an unmodified GRPO baseline in all eight (model, data seed) pairs we trained, by $+6.55$ points of macro avg@32 on average for MiniCPM5-2B-Base, mostly on problems of intermediate difficulty. The gain comes with longer answers, and across our runs accuracy and length move together (Section~\ref{sec:lenacc}). Three comparisons would separate where the credit goes from how long the model reasons: the entropy score of \citet{yeo2026surprising} under the same weighting, a random-score control together with a $\kappa=0$ run through the same trainer, and a length-matched GRPO baseline.

\section*{Limitations}
\begin{itemize}
  \item \textbf{No entropy baseline.} We compare only with GRPO. We did not run the entropy score of \citet{yeo2026surprising} or any other entropy-based credit, so we cannot say whether the margin beats entropy; we can say only that MCPO beats uniform credit.
  \item \textbf{Baseline strength.} We compare with one GRPO configuration (200 steps, LoRA, default learning rate). Two of its runs produce very short answers. A stronger GRPO baseline (longer training, tuned learning rate, or full fine-tuning) may narrow the gap.
  \item \textbf{Length.} MCPO models write 1.2 to 4.6 times longer answers than their GRPO counterparts, and the gain grows with this ratio (Section~\ref{sec:length}). Without a length-matched control or an ablation of the thresholds and $\kappa$, we cannot attribute the gain to the credit signal alone.
  \item \textbf{Scale.} We use two base models (2B and 4B parameters), LoRA, 200 steps, 2{,}048 training problems per seed, one task family (competition mathematics), and one untuned setting of $\kappa$ and the thresholds. With three seeds, the Qwen3-4B-Base gain does not reach significance under a $t$-test. We do not report scores of the base models before training.
  \item \textbf{Untested interactions and costs.} Token advantages enter the clipped loss directly, so MCPO may interact with the sampling, clipping and length settings we kept fixed. We did not measure wall-clock overhead, generated-token totals, or output diversity.
  \item \textbf{Approximations.} Margins are computed in bf16 and are one optimizer step old for the second mini-batch of a rollout. We tested the method only in single-process training.
  \item \textbf{Variance and contamination.} Data seeds change which training problems are drawn; the training seed and the evaluation sampling seed stay fixed. Our intervals therefore omit initialization and sampling noise. We did not check the training problems for overlap with the benchmarks beyond the codebase's own data pipeline; contamination is a known risk for Qwen base models in particular~\citep{wu2025reasoning}, although the two 2026 benchmarks postdate Qwen3-4B-Base and show gains of the same sign for every seed.
\end{itemize}

\section*{Ethics Statement}
This work changes how a language model is optimized on public mathematics data and evaluates it on public benchmarks. It involves no personal data and no human annotation. Stronger mathematical reasoning carries the usual dual-use concerns of general capability gains. Each training run used one 80\,GB GPU for 200 steps; each evaluation drew 32 samples for each of 297 problems. MCPO models write longer answers, which raises the inference cost per query.

\bibliography{references}

\begin{thebibliography}{64}
\providecommand{\natexlab}[1]{#1}

\bibitem[{Agarwal et~al.(2021)Agarwal, Schwarzer, Castro, Courville, and Bellemare}]{agarwal2021precipice}
Rishabh Agarwal, Max Schwarzer, Pablo~Samuel Castro, Aaron Courville, and Marc~G. Bellemare. 2021.
\newblock Deep reinforcement learning at the edge of the statistical precipice.
\newblock In \emph{Advances in Neural Information Processing Systems (NeurIPS)}.

\bibitem[{Agarwal et~al.(2025)Agarwal, Zhang, Yuan, Han, and Peng}]{agarwal2025entmin}
Shivam Agarwal, Zimin Zhang, Lifan Yuan, Jiawei Han, and Hao Peng. 2025.
\newblock The unreasonable effectiveness of entropy minimization in {LLM} reasoning.
\newblock \emph{arXiv preprint arXiv:2505.15134}.

\bibitem[{Aggarwal and Welleck(2025)}]{aggarwal2025l1}
Pranjal Aggarwal and Sean Welleck. 2025.
\newblock {L1}: Controlling how long a reasoning model thinks with reinforcement learning.
\newblock \emph{arXiv preprint arXiv:2503.04697}.

\bibitem[{Ahmadian et~al.(2024)Ahmadian, Cremer, Gallé, Fadaee, Kreutzer, Pietquin, Üstün, and Hooker}]{ahmadian2024rloo}
Arash Ahmadian, Chris Cremer, Matthias Gallé, Marzieh Fadaee, Julia Kreutzer, Olivier Pietquin, Ahmet Üstün, and Sara Hooker. 2024.
\newblock Back to basics: Revisiting reinforce style optimization for learning from human feedback in {LLMs}.
\newblock In \emph{Proceedings of the 62nd Annual Meeting of the Association for Computational Linguistics (ACL)}.

\bibitem[{Bigelow et~al.(2024)Bigelow, Holtzman, Tanaka, and Ullman}]{bigelow2024forking}
Eric Bigelow, Ari Holtzman, Hidenori Tanaka, and Tomer Ullman. 2024.
\newblock Forking paths in neural text generation.
\newblock \emph{arXiv preprint arXiv:2412.07961}.

\bibitem[{Chen et~al.(2021)Chen, Tworek, Jun, Yuan, Pinto, Kaplan, Edwards, Burda, Joseph, Brockman, Ray, Puri, Krueger, Petrov, Khlaaf, Sastry, Mishkin, Chan, Gray, Ryder, Pavlov, Power, Kaiser, Bavarian, Winter, Tillet, Such, Cummings, Plappert, Chantzis, Barnes, Herbert-Voss, Guss, Nichol, Paino, Tezak, Tang, Babuschkin, Balaji, Jain, Saunders, Hesse, Carr, Leike, Achiam, Misra, Morikawa, Radford, Knight, Brundage, Murati, Mayer, Welinder, McGrew, Amodei, McCandlish, Sutskever, and Zaremba}]{chen2021codex}
Mark Chen, Jerry Tworek, Heewoo Jun, Qiming Yuan, Henrique Ponde de~Oliveira Pinto, Jared Kaplan, Harri Edwards, Yuri Burda, Nicholas Joseph, Greg Brockman, Alex Ray, Raul Puri, Gretchen Krueger, Michael Petrov, Heidy Khlaaf, Girish Sastry, Pamela Mishkin, Brooke Chan, Scott Gray, and 39 others. 2021.
\newblock Evaluating large language models trained on code.
\newblock \emph{arXiv preprint arXiv:2107.03374}.

\bibitem[{Chen et~al.(2025)Chen, Qin, Wu, Ling, Ye, Zhao, and Shi}]{chen2025passk}
Zhipeng Chen, Xiaobo Qin, Youbin Wu, Yue Ling, Qinghao Ye, Wayne~Xin Zhao, and Guang Shi. 2025.
\newblock Pass@k training for adaptively balancing exploration and exploitation of large reasoning models.
\newblock \emph{arXiv preprint arXiv:2508.10751}.

\bibitem[{Cheng et~al.(2025)Cheng, Huang, Zhu, Dai, Zhao, Zhang, and Wei}]{cheng2025exploration}
Daixuan Cheng, Shaohan Huang, Xuekai Zhu, Bo~Dai, Wayne~Xin Zhao, Zhenliang Zhang, and Furu Wei. 2025.
\newblock Reasoning with exploration: An entropy perspective.
\newblock \emph{arXiv preprint arXiv:2506.14758}.

\bibitem[{Cui et~al.(2025{\natexlab{a}})Cui, Yuan, Wang, Wang, Zhang, Chen, Li, He, Fan, Yu, Xu, Chen, Yuan, Chen, Zhang, Lv, Wang, Yao, Han, Peng, Cheng, Liu, Sun, Zhou, and Ding}]{cui2025prime}
Ganqu Cui, Lifan Yuan, Zefan Wang, Hanbin Wang, Yuchen Zhang, Jiacheng Chen, Wendi Li, Bingxiang He, Yuchen Fan, Tianyu Yu, Qixin Xu, Weize Chen, Jiarui Yuan, Huayu Chen, Kaiyan Zhang, Xingtai Lv, Shuo Wang, Yuan Yao, Xu~Han, and 6 others. 2025{\natexlab{a}}.
\newblock Process reinforcement through implicit rewards.
\newblock \emph{arXiv preprint arXiv:2502.01456}.

\bibitem[{Cui et~al.(2025{\natexlab{b}})Cui, Zhang, Chen, Yuan, Wang, Zuo, Li, Fan, Chen, Chen, Liu, Peng, Bai, Ouyang, Cheng, Zhou, and Ding}]{cui2025entropy}
Ganqu Cui, Yuchen Zhang, Jiacheng Chen, Lifan Yuan, Zhi Wang, Yuxin Zuo, Haozhan Li, Yuchen Fan, Huayu Chen, Weize Chen, Zhiyuan Liu, Hao Peng, Lei Bai, Wanli Ouyang, Yu~Cheng, Bowen Zhou, and Ning Ding. 2025{\natexlab{b}}.
\newblock The entropy mechanism of reinforcement learning for reasoning language models.
\newblock \emph{arXiv preprint arXiv:2505.22617}.

\bibitem[{{DeepSeek-AI}(2025)}]{guo2025r1}
{DeepSeek-AI}. 2025.
\newblock Deepseek-r1: Incentivizing reasoning capability in llms via reinforcement learning.
\newblock \emph{Nature}.

\bibitem[{Fatemi et~al.(2025)Fatemi, Rafiee, Tang, and Talamadupula}]{fatemi2025concise}
Mehdi Fatemi, Banafsheh Rafiee, Mingjie Tang, and Kartik Talamadupula. 2025.
\newblock Concise reasoning via reinforcement learning.
\newblock \emph{arXiv preprint arXiv:2504.05185}.

\bibitem[{Fu et~al.(2025)Fu, Wang, Tian, and Zhao}]{fu2025deepconf}
Yichao Fu, Xuewei Wang, Yuandong Tian, and Jiawei Zhao. 2025.
\newblock Deep think with confidence.
\newblock \emph{arXiv preprint arXiv:2508.15260}.

\bibitem[{Guo et~al.(2017)Guo, Pleiss, Sun, and Weinberger}]{guo2017calibration}
Chuan Guo, Geoff Pleiss, Yu~Sun, and Kilian~Q. Weinberger. 2017.
\newblock On calibration of modern neural networks.
\newblock In \emph{International Conference on Machine Learning (ICML)}.

\bibitem[{Guo et~al.(2025)Guo, Xu, Liu, Ye, and Qiu}]{guo2025spo}
Yiran Guo, Lijie Xu, Jie Liu, Dan Ye, and Shuang Qiu. 2025.
\newblock Segment policy optimization: Effective segment-level credit assignment in {RL} for large language models.
\newblock \emph{arXiv preprint arXiv:2505.23564}.

\bibitem[{Henderson et~al.(2018)Henderson, Islam, Bachman, Pineau, Precup, and Meger}]{henderson2018matters}
Peter Henderson, Riashat Islam, Philip Bachman, Joelle Pineau, Doina Precup, and David Meger. 2018.
\newblock Deep reinforcement learning that matters.
\newblock In \emph{Proceedings of the AAAI Conference on Artificial Intelligence}.

\bibitem[{Hendrycks et~al.(2021)Hendrycks, Burns, Kadavath, Arora, Basart, Tang, Song, and Steinhardt}]{hendrycks2021math}
Dan Hendrycks, Collin Burns, Saurav Kadavath, Akul Arora, Steven Basart, Eric Tang, Dawn Song, and Jacob Steinhardt. 2021.
\newblock Measuring mathematical problem solving with the {MATH} dataset.
\newblock In \emph{Advances in Neural Information Processing Systems (NeurIPS) Datasets and Benchmarks Track}.

\bibitem[{Hochlehnert et~al.(2025)Hochlehnert, Bhatnagar, Udandarao, Albanie, Prabhu, and Bethge}]{hochlehnert2025sober}
Andreas Hochlehnert, Hardik Bhatnagar, Vishaal Udandarao, Samuel Albanie, Ameya Prabhu, and Matthias Bethge. 2025.
\newblock A sober look at progress in language model reasoning: Pitfalls and paths to reproducibility.
\newblock \emph{arXiv preprint arXiv:2504.07086}.

\bibitem[{Hu et~al.(2022)Hu, Shen, Wallis, Allen-Zhu, Li, Wang, Wang, and Chen}]{hu2022lora}
Edward~J. Hu, Yelong Shen, Phillip Wallis, Zeyuan Allen-Zhu, Yuanzhi Li, Shean Wang, Lu~Wang, and Weizhu Chen. 2022.
\newblock Lora: Low-rank adaptation of large language models.
\newblock In \emph{International Conference on Learning Representations (ICLR)}.

\bibitem[{Hu et~al.(2025{\natexlab{a}})Hu, Liu, Xu, and Shen}]{hu2025reinforcepp}
Jian Hu, Jason~Klein Liu, Haotian Xu, and Wei Shen. 2025{\natexlab{a}}.
\newblock {REINFORCE++}: Stabilizing critic-free policy optimization with global advantage normalization.
\newblock \emph{arXiv preprint arXiv:2501.03262}.

\bibitem[{Hu et~al.(2025{\natexlab{b}})Hu, Zhang, Han, Jiang, Zhang, and Shum}]{hu2025orz}
Jingcheng Hu, Yinmin Zhang, Qi~Han, Daxin Jiang, Xiangyu Zhang, and Heung-Yeung Shum. 2025{\natexlab{b}}.
\newblock {Open-Reasoner-Zero}: An open source approach to scaling up reinforcement learning on the base model.
\newblock \emph{arXiv preprint arXiv:2503.24290}.

\bibitem[{Hu et~al.(2024)Hu, Tu, Han, He, Cui, Long, Zheng, Fang, Huang, Zhao, Zhang, Thai, Zhang, Wang, Yao, Zhao, Zhou, Cai, Zhai, Ding, Jia, Zeng, Li, Liu, and Sun}]{hu2024minicpm}
Shengding Hu, Yuge Tu, Xu~Han, Chaoqun He, Ganqu Cui, Xiang Long, Zhi Zheng, Yewei Fang, Yuxiang Huang, Weilin Zhao, Xinrong Zhang, Zheng~Leng Thai, Kaihuo Zhang, Chongyi Wang, Yuan Yao, Chenyang Zhao, Jie Zhou, Jie Cai, Zhongwu Zhai, and 6 others. 2024.
\newblock {MiniCPM}: Unveiling the potential of small language models with scalable training strategies.
\newblock \emph{arXiv preprint arXiv:2404.06395}.

\bibitem[{{Hugging Face}(2025)}]{mathverify}
{Hugging Face}. 2025.
\newblock Math-verify: a mathematical expression evaluation system.
\newblock \url{https://github.com/huggingface/Math-Verify}.

\bibitem[{Kadavath et~al.(2022)Kadavath, Conerly, Askell, Henighan, Drain, Perez, Schiefer, Hatfield-Dodds, DasSarma, Tran-Johnson, Johnston, El-Showk, Jones, Elhage, Hume, Chen, Bai, Bowman, Fort, Ganguli, Hernandez, Jacobson, Kernion, Kravec, Lovitt, Ndousse, Olsson, Ringer, Amodei, Brown, Clark, Joseph, Mann, McCandlish, Olah, and Kaplan}]{kadavath2022know}
Saurav Kadavath, Tom Conerly, Amanda Askell, Tom Henighan, Dawn Drain, Ethan Perez, Nicholas Schiefer, Zac Hatfield-Dodds, Nova DasSarma, Eli Tran-Johnson, Scott Johnston, Sheer El-Showk, Andy Jones, Nelson Elhage, Tristan Hume, Anna Chen, Yuntao Bai, Sam Bowman, Stanislav Fort, and 17 others. 2022.
\newblock Language models (mostly) know what they know.
\newblock \emph{arXiv preprint arXiv:2207.05221}.

\bibitem[{Kang et~al.(2025)Kang, Zhao, and Song}]{kang2025selfcertainty}
Zhewei Kang, Xuandong Zhao, and Dawn Song. 2025.
\newblock Scalable best-of-{N} selection for large language models via self-certainty.
\newblock \emph{arXiv preprint arXiv:2502.18581}.

\bibitem[{Kazemnejad et~al.(2024)Kazemnejad, Aghajohari, Portelance, Sordoni, Reddy, Courville, and Roux}]{kazemnejad2024vineppo}
Amirhossein Kazemnejad, Milad Aghajohari, Eva Portelance, Alessandro Sordoni, Siva Reddy, Aaron Courville, and Nicolas~Le Roux. 2024.
\newblock Vineppo: Refining credit assignment in rl training of llms.
\newblock \emph{arXiv preprint arXiv:2410.01679}.

\bibitem[{{Kimi Team}(2025)}]{kimi2025k15}
{Kimi Team}. 2025.
\newblock {Kimi} k1.5: Scaling reinforcement learning with {LLMs}.
\newblock \emph{arXiv preprint arXiv:2501.12599}.

\bibitem[{Kuhn et~al.(2023)Kuhn, Gal, and Farquhar}]{kuhn2023semantic}
Lorenz Kuhn, Yarin Gal, and Sebastian Farquhar. 2023.
\newblock Semantic uncertainty: Linguistic invariances for uncertainty estimation in natural language generation.
\newblock In \emph{International Conference on Learning Representations (ICLR)}.

\bibitem[{Kwon et~al.(2023)Kwon, Li, Zhuang, Sheng, Zheng, Yu, Gonzalez, Zhang, and Stoica}]{kwon2023vllm}
Woosuk Kwon, Zhuohan Li, Siyuan Zhuang, Ying Sheng, Lianmin Zheng, Cody~Hao Yu, Joseph~E. Gonzalez, Hao Zhang, and Ion Stoica. 2023.
\newblock Efficient memory management for large language model serving with pagedattention.
\newblock In \emph{Proceedings of the 29th Symposium on Operating Systems Principles (SOSP)}.

\bibitem[{Lambert et~al.(2024)Lambert, Morrison, Pyatkin, Huang, Ivison, Brahman, Miranda, Liu, Dziri, Lyu, Gu, Malik, Graf, Hwang, Yang, Bras, Tafjord, Wilhelm, Soldaini, Smith, Wang, Dasigi, and Hajishirzi}]{lambert2024tulu}
Nathan Lambert, Jacob Morrison, Valentina Pyatkin, Shengyi Huang, Hamish Ivison, Faeze Brahman, Lester James~V. Miranda, Alisa Liu, Nouha Dziri, Shane Lyu, Yuling Gu, Saumya Malik, Victoria Graf, Jena~D. Hwang, Jiangjiang Yang, Ronan~Le Bras, Oyvind Tafjord, Chris Wilhelm, Luca Soldaini, and 4 others. 2024.
\newblock Tulu 3: Pushing frontiers in open language model post-training.
\newblock \emph{arXiv preprint arXiv:2411.15124}.

\bibitem[{Li et~al.(2024)Li, Xu, Zhang, Lin, Yu, Sun, and Luo}]{li2024remax}
Ziniu Li, Tian Xu, Yushun Zhang, Zhihang Lin, Yang Yu, Ruoyu Sun, and Zhi-Quan Luo. 2024.
\newblock {ReMax}: A simple, effective, and efficient reinforcement learning method for aligning large language models.
\newblock In \emph{International Conference on Machine Learning (ICML)}.

\bibitem[{Lightman et~al.(2024)Lightman, Kosaraju, Burda, Edwards, Baker, Lee, Leike, Schulman, Sutskever, and Cobbe}]{lightman2023verify}
Hunter Lightman, Vineet Kosaraju, Yura Burda, Harri Edwards, Bowen Baker, Teddy Lee, Jan Leike, John Schulman, Ilya Sutskever, and Karl Cobbe. 2024.
\newblock Let's verify step by step.
\newblock In \emph{International Conference on Learning Representations (ICLR)}.

\bibitem[{Lin et~al.(2024)Lin, Liang, Xu, Lin, Wang, Luo, Shi, Li, Yang, and Tu}]{lin2024critical}
Zicheng Lin, Tian Liang, Jiahao Xu, Qiuzhi Lin, Xing Wang, Ruilin Luo, Chufan Shi, Siheng Li, Yujiu Yang, and Zhaopeng Tu. 2024.
\newblock Critical tokens matter: Token-level contrastive estimation enhances {LLM}'s reasoning capability.
\newblock \emph{arXiv preprint arXiv:2411.19943}.

\bibitem[{Liu et~al.(2025{\natexlab{a}})Liu, Diao, Lu, Hu, Dong, Choi, Kautz, and Dong}]{liu2025prorl}
Mingjie Liu, Shizhe Diao, Ximing Lu, Jian Hu, Xin Dong, Yejin Choi, Jan Kautz, and Yi~Dong. 2025{\natexlab{a}}.
\newblock {ProRL}: Prolonged reinforcement learning expands reasoning boundaries in large language models.
\newblock \emph{arXiv preprint arXiv:2505.24864}.

\bibitem[{Liu et~al.(2025{\natexlab{b}})Liu, Wang, Zhang, Li, Zhou, Gai, and Pan}]{liu2025archer}
Runze Liu, Jiakang Wang, Fuzheng Zhang, Xiu Li, Guorui Zhou, Kun Gai, and Ling Pan. 2025{\natexlab{b}}.
\newblock Stabilizing knowledge, promoting reasoning: Dual-token constraints for rlvr.
\newblock \emph{arXiv preprint arXiv:2507.15778}.

\bibitem[{Liu et~al.(2025{\natexlab{c}})Liu, Chen, Li, Qi, Pang, Du, Lee, and Lin}]{liu2025drgrpo}
Zichen Liu, Changyu Chen, Wenjun Li, Penghui Qi, Tianyu Pang, Chao Du, Wee~Sun Lee, and Min Lin. 2025{\natexlab{c}}.
\newblock Understanding {R1-Zero}-like training: A critical perspective.
\newblock \emph{arXiv preprint arXiv:2503.20783}.

\bibitem[{Prabhudesai et~al.(2025)Prabhudesai, Chen, Ippoliti, Fragkiadaki, Liu, and Pathak}]{prabhudesai2025rent}
Mihir Prabhudesai, Lili Chen, Alex Ippoliti, Katerina Fragkiadaki, Hao Liu, and Deepak Pathak. 2025.
\newblock Maximizing confidence alone improves reasoning.
\newblock \emph{arXiv preprint arXiv:2505.22660}.

\bibitem[{Scheffer et~al.(2001)Scheffer, Decomain, and Wrobel}]{scheffer2001active}
Tobias Scheffer, Christian Decomain, and Stefan Wrobel. 2001.
\newblock Active hidden {M}arkov models for information extraction.
\newblock In \emph{Advances in Intelligent Data Analysis (IDA)}, pages 309--318.

\bibitem[{Schulman et~al.(2016)Schulman, Moritz, Levine, Jordan, and Abbeel}]{schulman2016gae}
John Schulman, Philipp Moritz, Sergey Levine, Michael Jordan, and Pieter Abbeel. 2016.
\newblock High-dimensional continuous control using generalized advantage estimation.
\newblock In \emph{International Conference on Learning Representations (ICLR)}.

\bibitem[{Schulman et~al.(2017)Schulman, Wolski, Dhariwal, Radford, and Klimov}]{schulman2017ppo}
John Schulman, Filip Wolski, Prafulla Dhariwal, Alec Radford, and Oleg Klimov. 2017.
\newblock Proximal policy optimization algorithms.
\newblock \emph{arXiv preprint arXiv:1707.06347}.

\bibitem[{Setlur et~al.(2024)Setlur, Nagpal, Fisch, Geng, Eisenstein, Agarwal, Agarwal, Berant, and Kumar}]{setlur2024rewarding}
Amrith Setlur, Chirag Nagpal, Adam Fisch, Xinyang Geng, Jacob Eisenstein, Rishabh Agarwal, Alekh Agarwal, Jonathan Berant, and Aviral Kumar. 2024.
\newblock Rewarding progress: Scaling automated process verifiers for {LLM} reasoning.
\newblock \emph{arXiv preprint arXiv:2410.08146}.

\bibitem[{Settles(2009)}]{settles2009active}
Burr Settles. 2009.
\newblock Active learning literature survey.
\newblock Technical Report 1648, University of Wisconsin--Madison, Department of Computer Sciences.

\bibitem[{Shao et~al.(2025)Shao, Li, Xin, Geng, Wang, Oh, Du, Lambert, Min, Krishna, Tsvetkov, Hajishirzi, Koh, and Zettlemoyer}]{shao2025spurious}
Rulin Shao, Shuyue~Stella Li, Rui Xin, Scott Geng, Yiping Wang, Sewoong Oh, Simon~Shaolei Du, Nathan Lambert, Sewon Min, Ranjay Krishna, Yulia Tsvetkov, Hannaneh Hajishirzi, Pang~Wei Koh, and Luke Zettlemoyer. 2025.
\newblock Spurious rewards: Rethinking training signals in {RLVR}.
\newblock \emph{arXiv preprint arXiv:2506.10947}.

\bibitem[{Shao et~al.(2024)Shao, Wang, Zhu, Xu, Song, Bi, Zhang, Zhang, Li, Wu, and Guo}]{shao2024deepseekmath}
Zhihong Shao, Peiyi Wang, Qihao Zhu, Runxin Xu, Junxiao Song, Xiao Bi, Haowei Zhang, Mingchuan Zhang, Y.~K. Li, Y.~Wu, and Daya Guo. 2024.
\newblock {DeepSeekMath}: Pushing the limits of mathematical reasoning in open language models.
\newblock \emph{arXiv preprint arXiv:2402.03300}.

\bibitem[{Sun et~al.(2025)Sun, Yang, Jian, Du, Cui, Ren, and Zhang}]{sun2025ktae}
Wei Sun, Wen Yang, Pu~Jian, Qianlong Du, Fuwei Cui, Shuo Ren, and Jiajun Zhang. 2025.
\newblock {KTAE}: A model-free algorithm to key-tokens advantage estimation in mathematical reasoning.
\newblock \emph{arXiv preprint arXiv:2505.16826}.

\bibitem[{Tan et~al.(2025)Tan, Wang, Pan, Lin, Wang, Wu, Chen, Zheng, Tang, and Yang}]{tan2025gtpo}
Hongze Tan, Zihan Wang, Jianfei Pan, Jinghao Lin, Hao Wang, Yifan Wu, Tao Chen, Zhihang Zheng, Zhihao Tang, and Haihua Yang. 2025.
\newblock Gtpo and grpo-s: Token and sequence-level reward shaping with policy entropy.
\newblock \emph{arXiv preprint arXiv:2508.04349}.

\bibitem[{Wang et~al.(2024)Wang, Li, Shao, Xu, Dai, Li, Chen, Wu, and Sui}]{wang2024mathshepherd}
Peiyi Wang, Lei Li, Zhihong Shao, R.~X. Xu, Damai Dai, Yifei Li, Deli Chen, Y.~Wu, and Zhifang Sui. 2024.
\newblock {Math-Shepherd}: Verify and reinforce {LLMs} step-by-step without human annotations.
\newblock In \emph{Proceedings of the 62nd Annual Meeting of the Association for Computational Linguistics (ACL)}.

\bibitem[{Wang et~al.(2025)Wang, Yu, Gao, Zheng, Liu, Lu, Dang, Chen, Yang, Zhang, Liu, Yang, Zhao, Yue, Song, Yu, Huang, and Lin}]{wang2025beyond}
Shenzhi Wang, Le~Yu, Chang Gao, Chujie Zheng, Shixuan Liu, Rui Lu, Kai Dang, Xionghui Chen, Jianxin Yang, Zhenru Zhang, Yuqiong Liu, An~Yang, Andrew Zhao, Yang Yue, Shiji Song, Bowen Yu, Gao Huang, and Junyang Lin. 2025.
\newblock Beyond the 80/20 rule: High-entropy minority tokens drive effective reinforcement learning for {LLM} reasoning.
\newblock \emph{arXiv preprint arXiv:2506.01939}.

\bibitem[{Wang et~al.(2023)Wang, Wei, Schuurmans, Le, Chi, Narang, Chowdhery, and Zhou}]{wang2023selfconsistency}
Xuezhi Wang, Jason Wei, Dale Schuurmans, Quoc Le, Ed~Chi, Sharan Narang, Aakanksha Chowdhery, and Denny Zhou. 2023.
\newblock Self-consistency improves chain of thought reasoning in language models.
\newblock In \emph{International Conference on Learning Representations (ICLR)}.

\bibitem[{Wang and Zhou(2024)}]{wang2024cotdecoding}
Xuezhi Wang and Denny Zhou. 2024.
\newblock Chain-of-thought reasoning without prompting.
\newblock In \emph{Advances in Neural Information Processing Systems (NeurIPS)}.

\bibitem[{Wei et~al.(2022)Wei, Wang, Schuurmans, Bosma, Ichter, Xia, Chi, Le, and Zhou}]{wei2022cot}
Jason Wei, Xuezhi Wang, Dale Schuurmans, Maarten Bosma, Brian Ichter, Fei Xia, Ed~Chi, Quoc Le, and Denny Zhou. 2022.
\newblock Chain-of-thought prompting elicits reasoning in large language models.
\newblock In \emph{Advances in Neural Information Processing Systems (NeurIPS)}.

\bibitem[{Wu et~al.(2025)Wu, Zhang, Dong, Xi, Zhao, Jin, Fan, Zhou, Lv, Zhang, Fu, Liu, Zhang, and Zhang}]{wu2025reasoning}
Mingqi Wu, Zhihao Zhang, Qiaole Dong, Zhiheng Xi, Jun Zhao, Senjie Jin, Xiaoran Fan, Yuhao Zhou, Huijie Lv, Ming Zhang, Yanwei Fu, Qin Liu, Songyang Zhang, and Qi~Zhang. 2025.
\newblock Reasoning or memorization? unreliable results of reinforcement learning due to data contamination.
\newblock \emph{arXiv preprint arXiv:2507.10532}.

\bibitem[{Yang et~al.(2025{\natexlab{a}})Yang, Li, Yang, Zhang, Hui, Zheng, Yu, Gao, Huang, Lv, Zheng, Liu, Zhou, Huang, Hu, Ge, Wei, Lin, Tang, Yang, Tu, Zhang, Yang, Yang, Zhou, Zhou, Lin, Dang, Bao, Yang, Yu, Deng, Li, Xue, Li, Zhang, Wang, Zhu, Men, Gao, Liu, Luo, Li, Tang, Yin, Ren, Wang, Zhang, Ren, Fan, Su, Zhang, Zhang, Wan, Liu, Wang, Cui, Zhang, Zhou, and Qiu}]{yang2025qwen3}
An~Yang, Anfeng Li, Baosong Yang, Beichen Zhang, Binyuan Hui, Bo~Zheng, Bowen Yu, Chang Gao, Chengen Huang, Chenxu Lv, Chujie Zheng, Dayiheng Liu, Fan Zhou, Fei Huang, Feng Hu, Hao Ge, Haoran Wei, Huan Lin, Jialong Tang, and 41 others. 2025{\natexlab{a}}.
\newblock {Qwen3} technical report.
\newblock \emph{arXiv preprint arXiv:2505.09388}.

\bibitem[{Yang et~al.(2024)Yang, Zhang, Hui, Gao, Yu, Li, Liu, Tu, Zhou, Lin, Lu, Xue, Lin, Liu, Ren, and Zhang}]{yang2024qwen25math}
An~Yang, Beichen Zhang, Binyuan Hui, Bofei Gao, Bowen Yu, Chengpeng Li, Dayiheng Liu, Jianhong Tu, Jingren Zhou, Junyang Lin, Keming Lu, Mingfeng Xue, Runji Lin, Tianyu Liu, Xingzhang Ren, and Zhenru Zhang. 2024.
\newblock {Qwen2.5-Math} technical report: Toward mathematical expert model via self-improvement.
\newblock \emph{arXiv preprint arXiv:2409.12122}.

\bibitem[{Yang et~al.(2025{\natexlab{b}})Yang, Luo, Wang, Han, He, Li, and Xu}]{yang2025lowprob}
Zhihe Yang, Xufang Luo, Zilong Wang, Dongqi Han, Zhiyuan He, Dongsheng Li, and Yunjian Xu. 2025{\natexlab{b}}.
\newblock Do not let low-probability tokens over-dominate in {RL} for {LLMs}.
\newblock \emph{arXiv preprint arXiv:2505.12929}.

\bibitem[{Yeo et~al.(2025)Yeo, Tong, Niu, Neubig, and Yue}]{yeo2025demystifying}
Edward Yeo, Yuxuan Tong, Morry Niu, Graham Neubig, and Xiang Yue. 2025.
\newblock Demystifying long chain-of-thought reasoning in {LLMs}.
\newblock \emph{arXiv preprint arXiv:2502.03373}.

\bibitem[{Yeo et~al.(2026)Yeo, Kang, Lee, Park, Baek, and Hwang}]{yeo2026surprising}
Woongyeong Yeo, Minki Kang, Chanuk Lee, Sangwoo Park, Jinheon Baek, and Sung~Ju Hwang. 2026.
\newblock Surprising success, repeated failure: Entropy-guided credit assignment for exploration in {LLM} reasoning.
\newblock \emph{arXiv preprint arXiv:2609.33781}.

\bibitem[{Yu et~al.(2025)Yu, Zhang, Zhu, Yuan, Zuo, Yue, Dai, Fan, Liu, Liu, Liu, Lin, Lin, Ma, Sheng, Tong, Zhang, Zhang, Zhang, Zhu, Zhu, Chen, Chen, Wang, Yu, Song, Wei, Zhou, Liu, Ma, Zhang, Yan, Qiao, Wu, and Wang}]{yu2025dapo}
Qiying Yu, Zheng Zhang, Ruofei Zhu, Yufeng Yuan, Xiaochen Zuo, Yu~Yue, Weinan Dai, Tiantian Fan, Gaohong Liu, Lingjun Liu, Xin Liu, Haibin Lin, Zhiqi Lin, Bole Ma, Guangming Sheng, Yuxuan Tong, Chi Zhang, Mofan Zhang, Wang Zhang, and 16 others. 2025.
\newblock {DAPO}: An open-source {LLM} reinforcement learning system at scale.
\newblock \emph{arXiv preprint arXiv:2503.14476}.

\bibitem[{Yue et~al.(2025{\natexlab{a}})Yue, Chen, Lu, Zhao, Wang, Yue, Song, and Huang}]{yue2025does}
Yang Yue, Zhiqi Chen, Rui Lu, Andrew Zhao, Zhaokai Wang, Yang Yue, Shiji Song, and Gao Huang. 2025{\natexlab{a}}.
\newblock Does reinforcement learning really incentivize reasoning capacity in {LLMs} beyond the base model?
\newblock \emph{arXiv preprint arXiv:2504.13837}.

\bibitem[{Yue et~al.(2025{\natexlab{b}})Yue, Yuan, Yu, Zuo, Zhu, Xu, Chen, Wang, Fan, Du, Wei, Yu, Liu, Liu, Liu, Lin, Lin, Ma, Zhang, Zhang, Zhang, Zhu, Zhang, Liu, Wang, Wu, and Yan}]{yue2025vapo}
Yu~Yue, Yufeng Yuan, Qiying Yu, Xiaochen Zuo, Ruofei Zhu, Wenyuan Xu, Jiaze Chen, Chengyi Wang, TianTian Fan, Zhengyin Du, Xiangpeng Wei, Xiangyu Yu, Gaohong Liu, Juncai Liu, Lingjun Liu, Haibin Lin, Zhiqi Lin, Bole Ma, Chi Zhang, and 8 others. 2025{\natexlab{b}}.
\newblock {VAPO}: Efficient and reliable reinforcement learning for advanced reasoning tasks.
\newblock \emph{arXiv preprint arXiv:2504.05118}.

\bibitem[{Zelikman et~al.(2022)Zelikman, Wu, Mu, and Goodman}]{zelikman2022star}
Eric Zelikman, Yuhuai Wu, Jesse Mu, and Noah~D. Goodman. 2022.
\newblock {STaR}: Bootstrapping reasoning with reasoning.
\newblock In \emph{Advances in Neural Information Processing Systems (NeurIPS)}.

\bibitem[{Zeng et~al.(2025)Zeng, Huang, Liu, Liu, He, Ma, and He}]{zeng2025simplerl}
Weihao Zeng, Yuzhen Huang, Qian Liu, Wei Liu, Keqing He, Zejun Ma, and Junxian He. 2025.
\newblock {SimpleRL-Zoo}: Investigating and taming zero reinforcement learning for open base models in the wild.
\newblock \emph{arXiv preprint arXiv:2503.18892}.

\bibitem[{Zhao et~al.(2025)Zhao, Kang, Feng, Levine, and Song}]{zhao2025intuitor}
Xuandong Zhao, Zhewei Kang, Aosong Feng, Sergey Levine, and Dawn Song. 2025.
\newblock Learning to reason without external rewards.
\newblock \emph{arXiv preprint arXiv:2505.19590}.

\bibitem[{Zheng et~al.(2025)Zheng, Liu, Li, Chen, Yu, Gao, Dang, Liu, Men, Yang, Zhou, and Lin}]{zheng2025gspo}
Chujie Zheng, Shixuan Liu, Mingze Li, Xiong-Hui Chen, Bowen Yu, Chang Gao, Kai Dang, Yuqiong Liu, Rui Men, An~Yang, Jingren Zhou, and Junyang Lin. 2025.
\newblock Group sequence policy optimization.
\newblock \emph{arXiv preprint arXiv:2507.18071}.

\end{thebibliography}

\appendix
\section{Per-seed, per-benchmark gains}
\label{app:bench}
Table~\ref{tab:perseed} gives the gain of MCPO over GRPO (avg@32, percentage points) for every data seed and benchmark.
\begin{table}[h]
\centering\footnotesize
\setlength{\tabcolsep}{2pt}
\resizebox{\columnwidth}{!}{\begin{tabular}{lrrrrrr}
\toprule
\textbf{Seed} & AIME24 & AIME25 & AIME26 & HMMT26 & AMC23 & M500\\
\midrule
\multicolumn{7}{l}{\emph{MiniCPM5-2B-Base}}\\
43 & +8.85 & +1.46 & +3.02 & +1.04 & +1.33 & +3.01\\
45 & +10.63 & +6.25 & +7.08 & +2.27 & +7.19 & +7.81\\
100 & +11.15 & +6.04 & +3.65 & +2.18 & +4.45 & +6.44\\
1024 & +12.60 & +6.46 & +9.69 & +3.79 & +5.00 & +7.91\\
31415 & +17.40 & +5.52 & +9.79 & +5.59 & +8.05 & +10.82\\
\midrule
\multicolumn{7}{l}{\emph{Qwen3-4B-Base}}\\
42 & +1.15 & +1.25 & +3.54 & +0.57 & +2.03 & +3.73\\
43 & +7.92 & +8.02 & +8.85 & +5.49 & +2.73 & +9.26\\
45 & +5.00 & +8.65 & +5.73 & +3.50 & +0.94 & +7.14\\
\bottomrule
\end{tabular}}
\caption{Gain (margin $-$ GRPO) per data seed and benchmark (M500 = MATH500 level 5). All 48 entries are positive.}
\label{tab:perseed}
\end{table}

\section{Accuracy--length correlations within a method}
\label{app:within}
Within a single method the number of runs is small, and the correlations between macro avg@32 and mean answer length are unstable: $0.89$ for MiniCPM GRPO and $0.34$ for MiniCPM MCPO (five runs each); $-0.52$ for Qwen3-4B GRPO and $0.95$ for Qwen3-4B MCPO (three runs each). Across seeds, MiniCPM MCPO scores vary less than GRPO scores (standard deviation 2.7 vs.\ 3.9 points); on Qwen3-4B the order is reversed (2.6 vs.\ 0.2).

\section{Training and evaluation settings}
Method hyperparameters: $\kappa=\ln4$, $\ell=0.1$, $u=0.7$. Training: LoRA $r=32$, $\alpha=64$, learning rate $10^{-5}$, 128 completions per rollout (16 prompts $\times$ 8), 200 steps, maximum completion length 10{,}240 tokens (soft overlong penalty starting at 8{,}192), vLLM GPU memory fraction 0.4. Evaluation: 297 problems, 32 samples, temperature 0.7, top-$p$ 0.8, top-$k$ 20, 16{,}384 new tokens. Software: torch 2.13.0, vLLM 0.30.0, TRL 1.14.0, transformers 5.17.0, PEFT 0.21.0, Math-Verify 0.9.0.

\end{document}